\documentclass[article, 10 pt, conference]{ieeeconf} 
\IEEEoverridecommandlockouts                          
\UseRawInputEncoding                                  
\let\labelindent\relax
\usepackage{enumitem}
\setlist{topsep=1pt, partopsep=0pt, parsep=0pt, itemsep=0pt}
\usepackage[
top    = 0.75in,
bottom = 0.75in,
left   = 0.75in,
right  = 0.75in]{geometry}
\usepackage{cite}
\usepackage[skip=10pt plus1pt, indent=35pt]{parskip}
\usepackage{placeins}
\usepackage{textcomp}
\usepackage[dvipsnames]{xcolor}
\definecolor{royalblue}{RGB}{65, 105, 225}
\definecolor{maroon}{RGB}{180, 0, 0}
\definecolor{DarkGreen}{RGB}{0, 100, 0}
\usepackage{booktabs}
\usepackage{multirow}
\usepackage{siunitx}
\usepackage{graphicx}
\usepackage[symbol]{footmisc}
\usepackage{adjustbox}
\usepackage{float}
\usepackage{svg}
\usepackage{array}
\usepackage[T1]{fontenc}
\usepackage{subcaption}
\usepackage{xspace}
\usepackage{scrextend}
\usepackage{caption}
\usepackage{tabularx}
\usepackage{listings}
\usepackage{wrapfig}
\usepackage{calligra}
\usepackage{comment}
\usepackage{soul}
\usepackage{balance}
\usepackage{listings}
\usepackage{xcolor}
\usepackage{caption}

\newcolumntype{A}{ >{\centering\arraybackslash} m{4cm} }
\newcolumntype{B}{ >{\centering\arraybackslash} m{1cm} }
\newcolumntype{C}[1]{>{\centering\let\newline\\\arraybackslash\hspace{0pt}}m{#1}}

\makeatletter
\newcommand\footnoteref[1]{\protected@xdef\@thefnmark{\ref{#1}}\@footnotemark}
\makeatother

\usepackage{amsmath,lipsum} 
\usepackage{amssymb} 
\usepackage{amsfonts}
\usepackage{textgreek}
\usepackage{authblk}
\usepackage{etoolbox}

\makeatletter
\let\NAT@parse\undefined

\let\oldthebibliography\thebibliography
\let\endoldthebibliography\endthebibliography
\renewenvironment{thebibliography}[1]{
  \oldthebibliography{#1}
  \setlength{\itemsep}{-1.75ex plus-.1ex minus-.1ex} 
}{
  \endoldthebibliography
}

\makeatother

\usepackage[colorlinks=true, citecolor=cyan]{hyperref}  
\def\BibTeX{{\rm B\kern-.05em{\sc i\kern-.025em b}\kern-.08em
    T\kern-.1667em\lower.7ex\hbox{E}\kern-.125emX}}

\usepackage{censor}
\usepackage{booktabs}
\usepackage{multirow}
\usepackage{caption}
\usepackage{geometry}
\usepackage[table]{xcolor}
\usepackage{booktabs}
\usepackage{multirow}
\usepackage{soul}

\definecolor{frontier}{RGB}{220,235,250}   
\definecolor{finetuned}{RGB}{220,245,220}  
\definecolor{base}{RGB}{250,220,220}       
\definecolor{additional}{RGB}{255,245,200} 
\definecolor{lightbrown}{RGB}{235,220,200}   
\definecolor{lightpurple}{RGB}{230,220,245}  
\usepackage{multirow}
\usepackage{url}
\title{\LARGE \bf
\textit{Pause and Think}: A Dataset and Benchmark for Video-Grounded Assistive Action Suggestion
}

\author{Shivam Singh$^{\ast\dagger\ddagger}$, Saptarshi Majumder$^{\circ\ddagger}$, Pratik Prabhanjan Brahma$^{\circ}$, Zicheng Liu$^{\circ}$ and Emad Barsoum$^{\circ}$
\thanks{$^{\ddagger}$ Primary authors.}
\thanks{$^{\dagger}$ IIIT Hyderabad, \quad $^{\circ}$ Advanced Micro Devices, Inc. (AMD)}
\thanks{$^{\ast}$Work done during Shivam Singh's internship at Advanced Micro Devices, Inc. (AMD).}
}

\makeatletter
\renewcommand{\@seccntformat}[1]{%
  \protect\csname the#1\endcsname\protect\quad%
}
\makeatother

\renewcommand{\thesection}{\arabic{section}}
\renewcommand{\thesubsection}{\thesection.\arabic{subsection}}
\renewcommand{\thesubsubsection}{\thesubsection.\arabic{subsubsection}}

\begin{document}

\maketitle
\thispagestyle{empty}
\pagestyle{empty}

\begin{abstract}
Recent Vision-Language Models (VLMs) struggle with grounded reasoning, temporal consistency, and context-aware planning in videos. We introduce \textit{pause-and-think-T}, a reasoning-centric training dataset that encourages models to pause, reason over visual evidence, and produce concise, actionable responses. The dataset promotes structured reasoning prior to answer generation, guiding models toward human-like, scene-grounded assistance. We fine-tune a compact 4B-parameter model and evaluate it on our \textit{pause-and-think-B} benchmark targeting contextual understanding and goal-planning tasks. The model achieves 58.0\% accuracy at 59$\times$ fewer parameters than Qwen3-VL-235B (58.9\%), matching GPT-5.2 on scene understanding and surpassing GPT-4o. Beyond our benchmark, it also shows strong out-of-distribution performance on EgoThink and TempCompass, with substantial gains in affordance, assistance, attribute recognition, situated reasoning, and temporal order, without benchmark-specific training. Our results indicate that targeted reasoning supervision enables compact models to deliver actionable, visually grounded guidance while generalizing beyond training data, without requiring large-scale model expansion. Code and data are available at \url{https://github.com/sssshivvvv/pause-and-think}.
\end{abstract}


\vspace{-1em}

\section{Introduction}

Consider a household scenario where a user needs assistance while performing multi-step tasks. They may ask: \textit{What is this object? Where are my keys? What should I do next?} An intelligent assistant that interprets a live video stream-captured from a first-person (e.g., wearable glasses) or third-person viewpoint-can provide grounded, concise guidance. For example: \textit{Pick up the screwdriver on your left, insert the screw you are holding, and tighten the wheel.} Such capabilities are valuable for cooking, assembly, and daily maintenance, particularly for users requiring contextual support.\\
Modern vision--language models (VLMs) exhibit strong perceptual and conversational abilities, yet often struggle with video-grounded actionable reasoning. Frontier models frequently produce verbose or generic responses that drift from visual evidence or hallucinate details. Existing video QA benchmarks primarily evaluate comprehension via multiple-choice selection. However, real assistive deployment requires concise, free-form instructions grounded in contextual understanding-not merely recognizing the correct option.\\
To bridge this gap, we introduce a reasoning-centric training dataset that encourages VLMs to \emph{think before suggesting}: a \textit{pause-and-think} paradigm in which the model deliberately pauses to reason over visual evidence before producing a concise, grounded response. The dataset promotes structured reasoning over temporally grounded video evidence, shifting model behavior from narration to concise, actionable, context-faithful assistance. It spans both egocentric and exocentric videos from diverse real-world activities, enabling generalization across perspectives.\\
We further evaluate models on a benchmark curated to test contextual understanding and goal planning in video-driven environments. A compact $\sim$4B-parameter model fine-tuned on our dataset achieves performance competitive with frontier systems while remaining efficient and edge-deployable.

Our contributions are:
\begin{itemize}[noitemsep, topsep=0pt]

    \item We introduce two reasoning-centric datasets: \textit{Pause-and-think-T} , a training set of $\sim$10k high-quality samples with structured reasoning supervision, and \textit{Pause-and-think-B}, a benchmark for evaluating assistive action suggestion that moves beyond recognition-based multiple-choice towards free-form actionable guidance.
    \item We show that targeted reasoning supervision on only \textit{Pause-and-think-T} enables a compact 4B model to achieve Pareto-efficient performance against frontier models.
    \item We demonstrate that this paradigm is more assistive for the user in tasks, providing grounded, concise, and context-faithful next-step instructions while significantly reducing the contextual drift and verbosity typical of larger models.
\end{itemize}

Quantitatively, our fine-tuned model achieves 58.0\% accuracy at 59$\times$ fewer parameters than Qwen3-VL-235B (58.9\%), matching GPT-5.2 on scene understanding while remaining suitable for real-time edge deployment.

\section{Related Work}
\label{sec:related_work}

\subsection{Multimodal Video Understanding and Temporal Grounding}
Recent benchmarks (Video-MME \cite{video_mme}, EgoSchema \cite{egoschema}, TempCompass \cite{tempcompass}, NExT-QA \cite{next_qa}) have revealed that contemporary Video-LLMs frequently over-index on static visual priors and fail to grasp genuine temporal causality or fine-grained sequential order. Models often perform comparably even when input frames are artificially shuffled, indicating reliance on appearance rather than temporal reasoning. These findings expose a fundamental gap: strong perceptual accuracy does not imply strong \emph{actionable} reasoning. Identifying an object is not the same as knowing what to do with it next. This motivates paradigms---like ours---that enforce strict temporal grounding through structured reasoning supervision rather than surface-level descriptive heuristics.

\subsection{Egocentric Vision, Spatial Intelligence, and Embodied Agent Planning}
The transition from exocentric to egocentric video understanding is a critical prerequisite for Embodied Artificial Intelligence. Egocentric perception presents unique challenges due to constant camera motion, severe physical occlusions, and the necessity to infer continuous human intent \cite{Damen2018EPICKITCHENS, grauman2022ego4dworld3000hours}. Benchmarks such as EgoThink and VidEgoThink \cite{egothink} have systematically mapped the dimensions of first-person intelligence, while wearable datasets like WAGIBench \cite{wagibench} highlight a massive performance gap between human predictability and model inference regarding user goals. Frontier systems such as Gemini Robotics ER \cite{gemini_25} are evaluated on extensive spatial benchmarks that test pixel-level localization, 3D cognitive mapping, and adversarial visual reasoning. However, these benchmarks predominantly rely on rigid multiple-choice or exact-match formats and do not evaluate free-form, context-aware assistance. While prior egocentric benchmarks test \emph{what a model perceives}, ours tests \emph{what a model can help a user do}. Our \textit{pause-and-think-B} benchmark occupies a distinct niche: evaluating generative, actionable video-grounded assistance, requiring models to synthesize temporal context and output concise, human-like next-step instructions.

\subsection{Deliberate Reasoning and Structured Visual Logic}
To address the pervasive issues of visual hallucination and contextual drift, recent research has shifted toward instantiating System-2 cognitive processing within multimodal architectures. Although text-centric intermediate logical generation has shown promise, its direct application to visual domains often results in causal misalignment, where the generated text contradicts observable evidence. To enforce visual faithfulness, reinforcement learning approaches such as VLM-R1 and Video-R1 \cite{video_r1, vlm_r1}, as well as models such as Cosmos-Reason1\cite{cosmos}, utilize structured policy optimization to autonomously incentivize logical adherence. Similarly, frontier systems like Gemini Robotics ER leverage latent ``thinking'' phases to orchestrate complex robotic actuation.
Unlike these approaches---which rely on RL-based reward signals, massive parameter counts, or implicit latent reasoning---our method takes a fundamentally different path: injecting explicit \textit{pause-and-think} supervision during fine-tuning of compact models (4B parameters). While RL methods learn \emph{what} to reason about through trial-and-error, our structured supervision prescribes \emph{how} to reason through explicit intermediate traces, compelling models to evaluate visible temporal evidence \emph{before} formulating suggestions. This enables edge-deployable models to deliver highly grounded, actionable guidance without RL infrastructure or frontier-scale compute.
\section{Problem Formulation and Framework}

\begin{figure*}[t]
\centering
\captionsetup{font=scriptsize}

\begin{subfigure}{\textwidth}
    \centering
    \includegraphics[width=\textwidth]{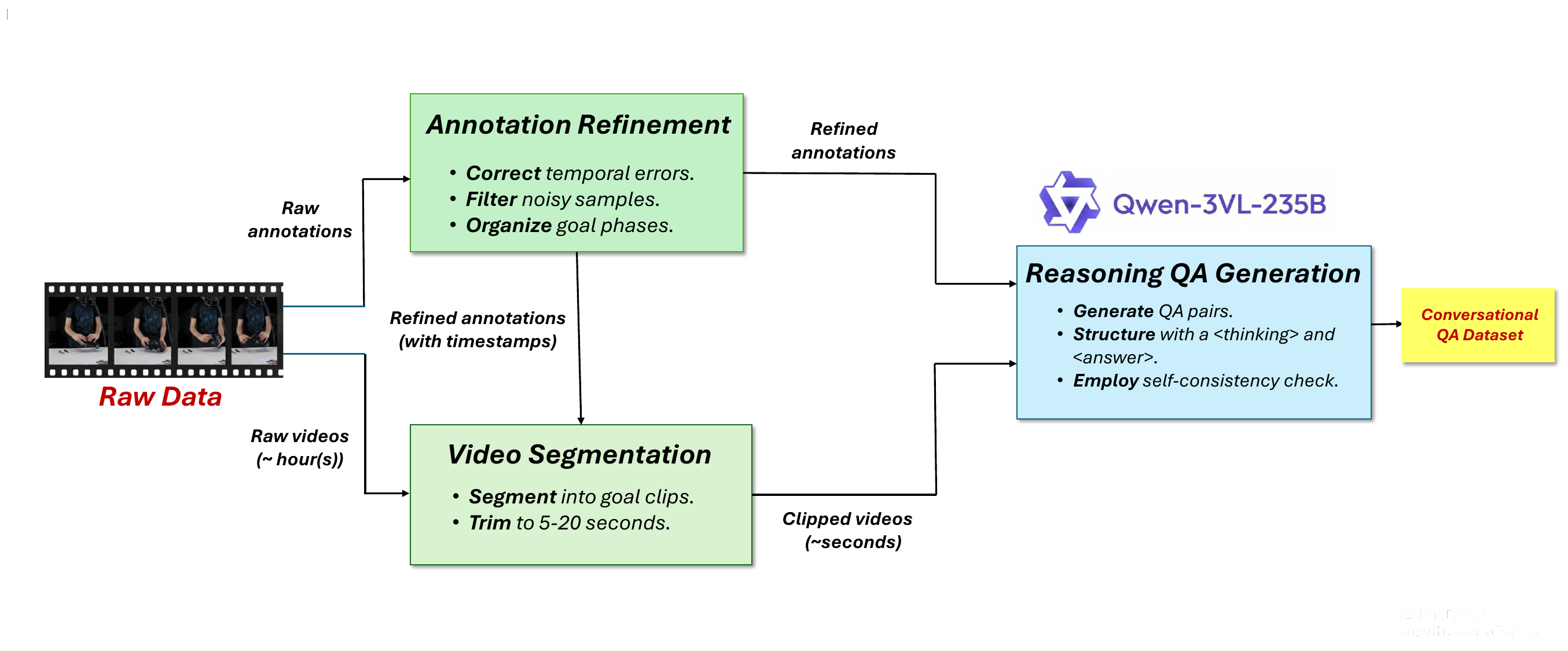}
\end{subfigure}

\vspace{-2em}

\begin{subfigure}{0.75\textwidth}
    \centering
    \includegraphics[width=\textwidth]{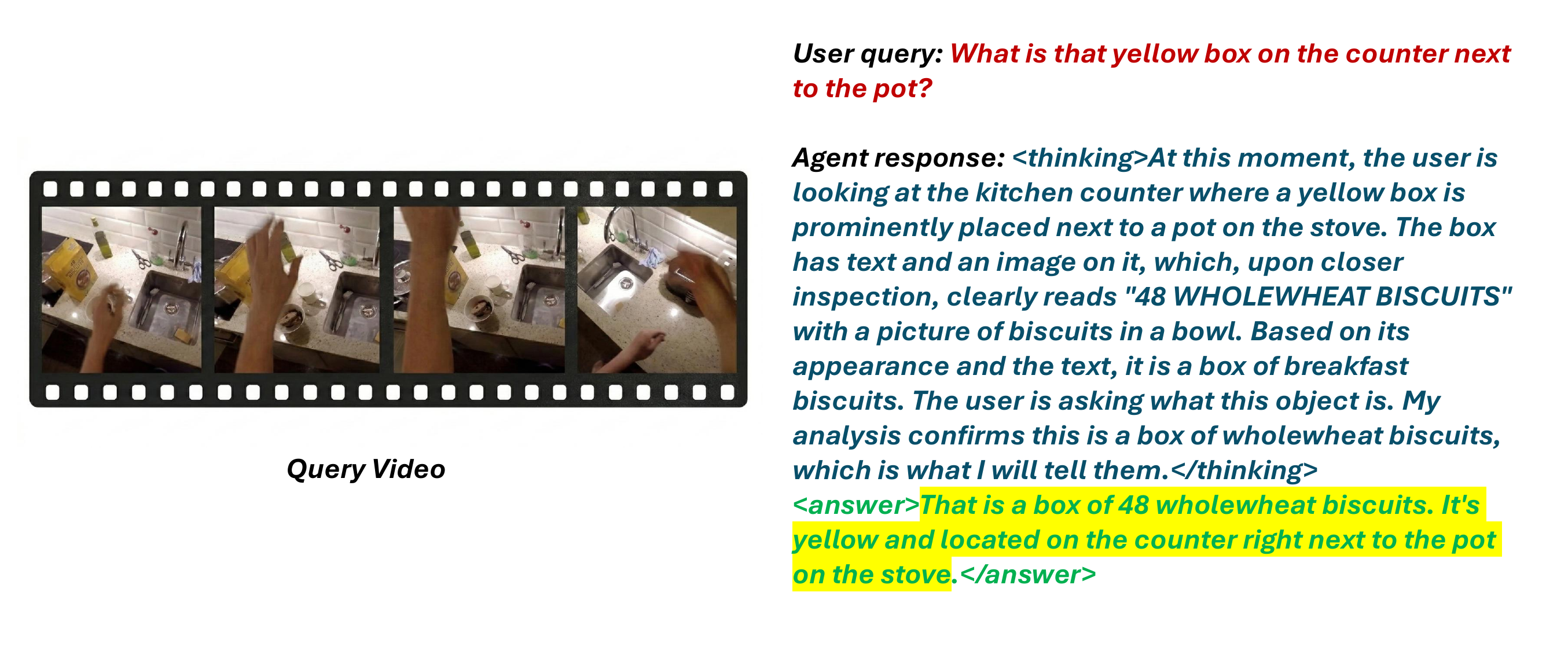}
\end{subfigure}

\caption{(Top) Dataset construction pipeline. Raw videos and annotations are refined using gpt-oss-120b to correct temporal inconsistencies, remove noise, and organize fine-grained actions into goal-oriented phases. Videos are segmented into short clips (5--20s), and Qwen-3VL-235B generates structured conversational QA pairs with explicit \texttt{<thinking>} and \texttt{<answer>} components, followed by self-consistency checks. This pipeline was used to build both the \textbf{training data and the proposed benchmark}, yielding a temporally grounded, reasoning-centric QA dataset. (Bottom) Example of a Conversational QA pair generated by the data pipeline shown above.}
\label{fig:pipeline_example}
\vspace{-0.5em}
\end{figure*}

\begin{figure}[tb]
\centering
\captionsetup{font=scriptsize}
\setlength{\belowcaptionskip}{-10pt}
\includegraphics[width=0.45\textwidth]{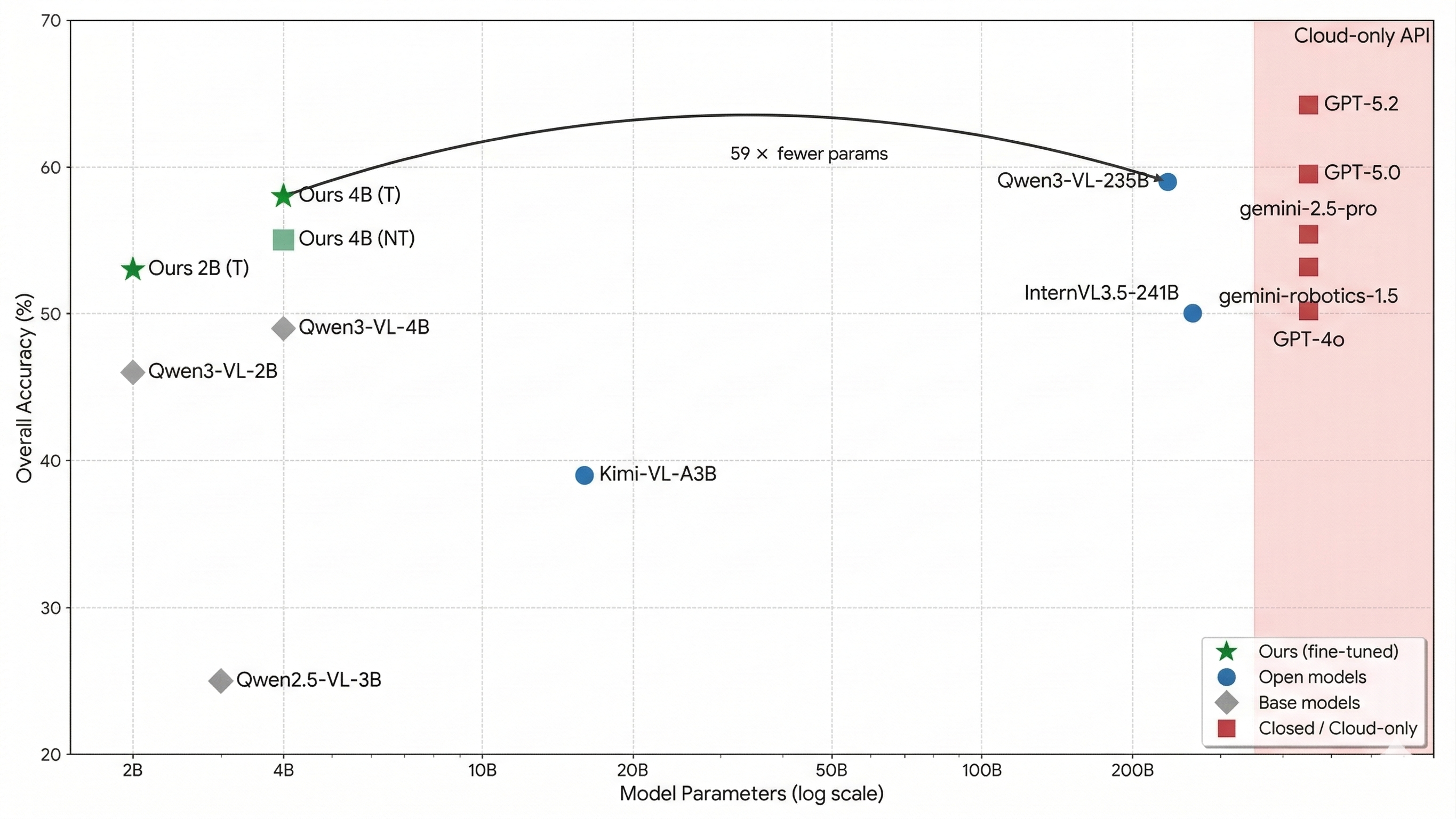}
\caption{ Accuracy vs.\ model scale on \textbf{our benchmark}. The fine-tuned 4B model lies on the open-weight Pareto frontier, achieving 58.0\% accuracy at 59$\times$ fewer parameters than Qwen3-VL-235B (58.9\%), while closed frontier models require cloud-only API access, limiting edge deployment.}
\vspace{-3pt}
\label{fig:pareto}
\end{figure}

\vspace{1em}

We study video-grounded agentic assistance, where a system interprets a user query within a dynamic visual scene and produces a concise, actionable response. The input is a short video segment and a natural language query expressing informational or goal-oriented intent. The objective is to generate a response that is temporally grounded and faithful to the visual evidence. Unlike video captioning or open-ended dialogue, this task requires actionable reasoning, including contextual understanding, object identification, and next-step planning. The system must bridge perception and decision-making by reasoning over visual cues before producing its final answer.

\subsection{Task Formulation}

Let $V = \{f_1, f_2, \dots, f_T\}$ denote a temporally ordered sequence of video frames and $q$ represent a user query. The goal is to learn a function:
\[
\mathcal{F}: (V, q) \rightarrow a
\]
where $a$ is a grounded, natural language response describing either contextual information about the scene or an actionable plan. To encourage deliberate reasoning, we decompose the response into two conceptual stages:
\[
(V, q) \rightarrow r \rightarrow a
\]
where $r$ represents intermediate reasoning grounded in visual evidence, and $a$ is the final concise answer presented to the user. Training supervision encourages the model to internally structure inference around this reasoning stage, promoting context-aware responses.

We consider two primary task categories:

\begin{itemize}
    \item \textbf{Contextual Question Answering:} Queries that require identifying objects, scene attributes, or ongoing actions.
    \item \textbf{Goal-Oriented Planning:} Queries that require predicting or suggesting the next action(s) consistent with the observed task progression.
\end{itemize}

\subsection{Data Construction Framework}

This section outlines the framework for transforming raw video datasets into reasoning-structured conversational QA data. As summarized in Figure~\ref{fig:pipeline_example}, the pipeline refines annotations, segments videos, and generates structured supervision while preserving temporal grounding and reducing noise. Long-form videos are converted into short, goal-oriented clips paired with conversational QA, through three main stages:

\begin{enumerate}
    \item Refinement of raw annotations to correct temporal inconsistencies and recover missing or ambiguous metadata.
    \item Goal-oriented video segmentation to produce short, context-preserving clips.
    \item Ground-truth QA generation with reasoning supervision and self-consistency validation.
\end{enumerate}

This staged pipeline ensures that the resulting dataset is temporally coherent, visually grounded, and aligned with actionable reasoning tasks.


\subsubsection{Annotation Refinement Framework}
\label{data_ref}
We construct the dataset from three large-scale video sources spanning egocentric and exocentric perspectives. Raw annotations frequently contain timestamp misalignment, missing labels, redundancy, and occlusion-induced ambiguity. To address this, we apply a unified refinement pipeline guided by a multimodal reasoning model. Fine-grained actions are temporally reordered and cleaned to remove noisy or incomplete segments, then grouped into coherent high-level goal phases to enable structured task progression. When coarse goals are absent, they are inferred from action sequences. We additionally extract contextual metadata (objects, verbs, interaction cues) to support QA generation. Refinement is performed using gpt-oss-120b~\cite{gpt-oss}, which produces temporally structured action lists and summaries aligned with the video content, ensuring semantic consistency while preserving visual grounding. Note that, \textit{ONLY} raw text annotations and their timestamps were given to gpt-oss during the refinement process; no video was given. We further filter videos with severe occlusions, incomplete annotations, redundant labeling, or weak interaction context, prioritizing clips with a single active user. The final annotations form temporally ordered, goal-aware metadata suitable for clip-level reasoning.

\subsubsection{Goal-Oriented Video Segmentation}


Long-form videos are segmented into compact clips capturing meaningful task transitions while maintaining visual coherence, using refined high-level goal phases as anchors. Clips are trimmed to 5--20 seconds. For planning scenarios, they are further split into prefix-suffix pairs representing query and continuation segments, enabling temporal reasoning over partial observations. Segmentation preserves interaction context while reducing redundancy; clips with heavy occlusion or low visual relevance are removed. The final set comprises diverse, temporally localized household and egocentric task scenarios.

\subsubsection{Ground-Truth QA Generation}

Ground-truth QA pairs are generated from refined annotations, goal metadata, and aligned video clips. For each clip, a user-style question is synthesized to reflect realistic assistive queries such as object identification or next-step planning.

We support two paradigms:
\begin{enumerate}
    \item \textbf{Contextual QA:} The entire clip serves as the observation window, targeting scene understanding, affordances, or task state.
    \item \textbf{Goal-Planning QA:} Each clip is split into a query and continuation segment. The final frame of the query defines the decision point; answers are validated against the continuation to ensure feasibility.
\end{enumerate}

Responses follow a structured format: \texttt{<thinking>} for intermediate reasoning and \texttt{<answer>} for the user-facing reply.
QA generation uses Qwen3-VL-235B-Instruct~\cite{Qwen3-VL}. To improve reliability, multiple candidates are generated per clip and majority voting across the generated candidates is used to select the final answer \cite{sc}. The selected response is then verified against annotations and video evidence and retained only if consistency checks pass. Sampling is controlled to balance diversity and correctness. The resulting dataset is temporally grounded and semantically verified for actionable task understanding. We train models with and without the thinking component, showing that structured reasoning improves inference quality and task success








\subsection{Training and Evaluation Protocol}
Our dataset is formatted as multimodal conversational supervision: each instance contains a video clip, a user query, and a reasoning-structured response. During fine-tuning, the model learns to associate visual evidence with intermediate reasoning patterns and concise final outputs. This setup encourages an implicit \emph{pause-and-think} behavior-grounding the scene before generating actionable responses-without exposing reasoning at inference time.\\
We evaluate models on our curated \textit{pause-and-think-B} benchmark targeting contextual understanding and goal planning in video-driven environments. Responses are compared against ground truth while verifying visual consistency. Evaluation emphasizes (i) grounding correctness, (ii) task relevance of suggested actions, and (iii) avoidance of hallucinations. For planning scenarios, multiple solutions are accepted if they remain context-consistent. This protocol measures whether models translate visual understanding into actionable assistance rather than descriptive narration.

\begin{figure*}[h]
  \centering
  \captionsetup{font=scriptsize}
  \includegraphics[width=\textwidth]{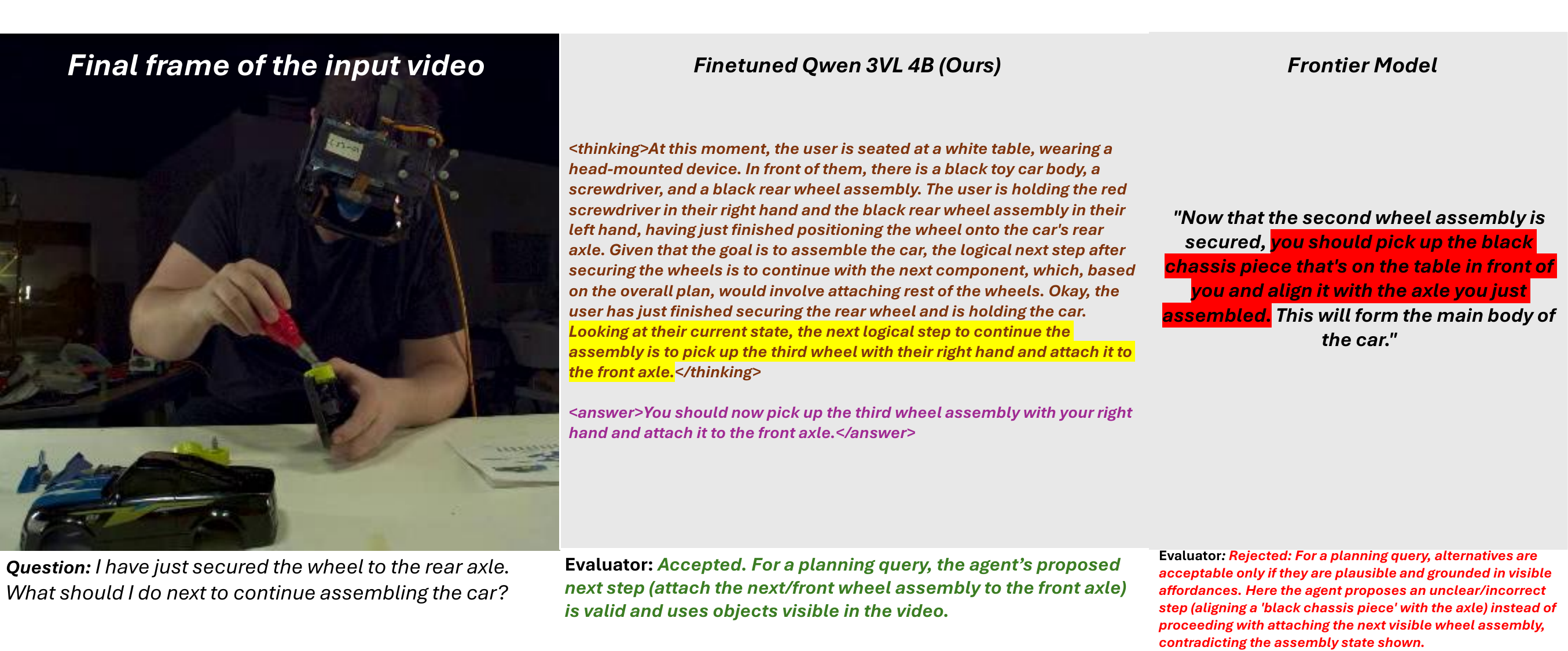}
  \vspace{-1em}
  \caption{Qualitative comparison between a 4B model fine-tuned on our \textit{pause-and-think-T} dataset and a frontier-scale model on a goal-planning query. Given the final video frame (left), the fine-tuned model (middle) produces structured \textit{pause-and-think} reasoning and correctly proposes attaching the third wheel assembly to the front axle, grounding its decision in visible evidence. In contrast, the frontier model (right) suggests attaching the chassis, overlooking the intermediate step of securing all wheels and exhibiting contextual drift. The evaluator accepts the fine-tuned response and rejects the frontier output, illustrating the grounding and planning advantages of reasoning-centric fine-tuning.}
  \label{fig:example}
  \vspace{-1em}
\end{figure*}


\section{Experimental Sections and Results}
\label{hypotheses}






\noindent
We evaluate three hypotheses on reasoning-centric supervision for video-grounded VLMs:

\vspace{-0.75em}
\begin{itemize}
\item[\textbf{H1:}] Reasoning-centric fine-tuning improves video-grounded understanding and increases actionable suggestion accuracy.

\item[\textbf{H2:}] Fine-tuning a compact 4B model on a high-quality small dataset with video-grounded reasoning enables performance comparable to frontier-scale systems in contextual understanding and action planning, without parameter scaling.

\item[\textbf{H3:}] Structured reasoning supervision reduces contextual drift and verbosity, enabling concise responses while maintaining a polite and user-friendly interaction style.
\end{itemize}

We test these on our curated \textit{pause-and-think-B} benchmark spanning scene understanding and goal planning tasks from egocentric and exocentric videos. We compare baseline, frontier, and fine-tuned models, focusing on visual grounding accuracy and action relevance to analyze the impact of reasoning-centric training on assistive performance.

\subsection{Experimental Setup}
\label{sec:expres-setup}


\subsubsection{Dataset Configuration}
Our training corpus is built from three large-scale video datasets: \textit{Epic-Kitchens}~\cite{Damen2018EPICKITCHENS}, \textit{Assembly101}~\cite{sener2022assembly101}, and \textit{Ego4D}~\cite{grauman2022ego4dworld3000hours}, selected for diverse perspectives and task contexts. After refinement (Section~\ref{data_ref}), the dataset contains 10,051 QA instances: 4,977 scene-understanding (49.5\%) and 5,074 goal-planning (50.5\%), yielding a near-balanced perception-planning split. \textit{Epic-Kitchens} contributes 5,873 instances (58.4\%), evenly split between 2,935 scene (49.9\%) and 2,938 goal examples (50.1\%). \textit{Assembly101} adds 2,058 instances (20.5\%), with 890 scene (43.2\%) and 1,168 goal examples (56.8\%). \textit{Ego4D} provides 2,120 instances (21.1\%), including 1,152 scene (54.3\%) and 968 goal examples (45.7\%). This composition ensures diversity across egocentric and exocentric views while maintaining balanced supervision for scene interpretation and forward planning.

\vspace{1em}
\subsubsection{Training}

We adopt Qwen3-VL-4B-Instruct as our backbone. The final training dataset, \textit{Pause-and-think-T},  consists of 10{,}051 curated video clips paired with reasoning-structured conversational supervision. All samples are formatted using the alpaca-style schema\cite{alpaca} to enable multimodal dialogue learning. Training was performed using the LLaMA-Factory framework\cite{zheng2024llamafactory} on a single node of 8$\times$ AMD Instinct\textsuperscript{TM} MI325 GPUs.
This setup enables efficient fine-tuning while preserving the reasoning-centric supervision necessary for video-grounded assistance tasks. We later performed inference using vLLM\cite{kwon2023efficient}.

\vspace{1em}
\subsubsection{Benchmark Data}
We construct a 300-sample video-grounded QA benchmark, \textit{Pause-and-think-B}, with 100 samples each from EPIC-Kitchens, Assembly101, and Ego4D to evaluate contextual scene understanding and goal planning. Candidate samples were screened using GPT-5 \cite{gpt}, and failure cases were selected to create a challenging subset (210 samples). An additional 90 samples were added for balance and coverage. The final benchmark supports systematic comparison across baseline, frontier, and fine-tuned models.

Across the pipeline, distinct models serve specialized roles: gpt-oss-120b refines annotations, Qwen3-VL-235B generates QA pairs, and GPT-5 screens benchmark difficulty. Given the subjective nature of action suggestion, we forgo a fixed-protocol human study and instead rely on qualitative analysis from independent human reviewers, who confirmed sample correctness on randomly selected QAs and found the resulting assistive execution plans to be well-grounded and more helpful for the user.

\vspace{1em}
\subsubsection{Evaluation Setup}

Our evaluation protocol directly tests the hypotheses in Sec.~\ref{hypotheses} by measuring grounded reasoning, action validity, and contextual alignment. We use GPT-5.1 \cite{gpt} as an automated multimodal evaluator under controlled criteria. Among several candidate evaluator models, our qualitative analysis found GPT-5.1 to be the most consistent judge, and its decisions aligned most closely with human evaluations on a randomly selected common set of QAs, motivating its use for at-scale evaluation.

Each evaluation instance is defined as
\[
\mathcal{E} = (V, q, a_{gt}, a_{m}),
\]
where \(V\) is the video, \(q\) the query, \(a_{gt}\) the ground truth, and \(a_{m}\) the model response. The evaluator receives the video, question, model answer, and ground truth, and explicitly verifies object presence, scene state, and action feasibility from visual evidence. For factual queries, correctness requires semantic agreement with \(a_{gt}\). For goal-planning tasks, visually grounded alternative plans that achieve the objective are accepted.

\noindent We employ a binary scoring strategy:

\begin{enumerate}
    \item[] \textbf{Binary Validity Scoring.}  
    The evaluator assigns a binary score
    \[
    S_{\text{bin}} \in \{0,1\},
    \]
is assigned, indicating whether the response is correct and visually grounded. Hallucinations, infeasible actions, or contextual contradictions result in rejection.
\end{enumerate}

All models are evaluated using the same prompt to ensure fairness. For thinking variants, only the \texttt{<answer>} is scored; the \texttt{<thinking>} segment is treated as internal reasoning and excluded from evaluation.

\begin{table*}[t]
\centering
\captionsetup{font=small, labelfont=bf}
\setlength{\tabcolsep}{8pt}

\begin{tabular}{p{6cm}ccc}
\toprule
\multirow{2}{*}{\textbf{Model Name}} &
\textbf{Overall (\%)} &
\textbf{Scene (\%)} &
\textbf{Goal (\%)} \\
& \textbf{NT / T} & \textbf{NT / T} & \textbf{NT / T} \\
\midrule
\noalign{\vskip -3pt}
\rowcolor{frontier}
\multicolumn{4}{l}{\textbf{Frontier Models (Closed)}} \\

\rowcolor{frontier}
\hspace{1em} 
GPT-5.2 & \textbf{64.24 / --} & \textbf{55.13 / --} & \textbf{86.52 / --} \\
\rowcolor{frontier}
\hspace{1em} GPT-5.0 & 59.27 / -- & 52.63 / -- & 74.19 / -- \\
\rowcolor{frontier}
\hspace{1em} Gemini-2.5-pro & 55.39 / -- & 49.05 / -- & 69.85 / -- \\
\rowcolor{frontier}
\hspace{1em} Gemini-robotics-ER-1.5-preview & 53.17 / -- & 45.24 / -- & 71.68 / -- \\
\rowcolor{frontier}
\hspace{1em} GPT-4o & 50.33 / -- & 42.58 / -- & 68.13 / -- \\
\rowcolor{frontier}
& & & \\

\rowcolor{lightpurple}
\multicolumn{4}{l}{\textbf{Frontier Models (Open)}} \\
\rowcolor{lightpurple}
\hspace{1em} Qwen-3VL-235B-Instruct & 58.89 / -- & 53.03 / -- & 72.26 / -- \\
\rowcolor{lightpurple}
& & & \\

\rowcolor{finetuned}
\multicolumn{4}{l}{\textbf{Finetuned Models (Frozen ViT and projector)}} \\
\rowcolor{finetuned}
\hspace{1em} Qwen-3VL-4B-Instruct \textit{(Ours)} & 54.67 / \textbf{58.00} & \textbf{55.77} / 55.02 & 52.17 / 64.84 \\
\rowcolor{finetuned}
\hspace{1em} Qwen-3VL-2B-Instruct & 51.33 / 52.67 & 50.00 / 52.38 & 54.44 / 53.33 \\
\rowcolor{finetuned}
\hspace{1em} Qwen-2.5VL-3B-Instruct & 36.11 / -- & 31.84 / -- & 46.58 / -- \\

\rowcolor{finetuned}
\multicolumn{4}{l}{\textbf{Finetuned Models (Full)}} \\
\rowcolor{finetuned}
\hspace{1em} Qwen-3VL-4B-Instruct & -- / 56.33 & -- / 51.67 & -- / \textbf{67.03} \\
\rowcolor{finetuned}
& & & \\

\rowcolor{base}
\multicolumn{4}{l}{\textbf{Base Models}} \\
\rowcolor{base}
\hspace{1em} Qwen-3VL-4B-Instruct & \textbf{49.00} / -- & \textbf{46.86} / -- & \textbf{53.76} / -- \\
\rowcolor{base}
\hspace{1em} Qwen-3VL-2B-Instruct & 45.67 / -- & 46.19 / -- & 44.44 / -- \\
\rowcolor{base}
\hspace{1em} Qwen-2.5VL-3B-Instruct & 24.60 / -- & 18.99 / -- & 38.36 / -- \\
\rowcolor{base}
& & & \\

\rowcolor{additional}
\multicolumn{4}{l}{\textbf{Additional Models}} \\
\rowcolor{additional}
\hspace{1em} InternVL3\_5-241B & \textbf{50.67} / -- & \textbf{43.81} / -- & \textbf{66.67} / -- \\
\rowcolor{additional}
\hspace{1em} Kimi-VL-A3B-Instruct & 39.33 / -- & 36.36 / -- & 46.15 / -- \\
\rowcolor{additional}
& & & \\

\rowcolor{finetuned}
\multicolumn{4}{l}{\textbf{Finetuned Models (Frozen ViT and projector)}} \\
\rowcolor{finetuned}
\hspace{1em} Qwen-3VL-4B-Thinking & \textbf{52.33} / -- & \textbf{50.95} / -- & \textbf{55.56} / -- \\
\rowcolor{finetuned}
\hspace{1em} Qwen-3VL-2B-Thinking & 44.67 / -- & 48.10 / -- & 36.67 / -- \\

\rowcolor{finetuned}
\multicolumn{4}{l}{\textbf{Finetuned Models (Full)}} \\
\rowcolor{finetuned}
\hspace{1em} Qwen-3VL-4B-Thinking & 52.00 / -- & 50.48 / -- & 55.56 / -- \\
\rowcolor{finetuned}
& & & \\

\noalign{\vskip -3pt}
\bottomrule
\end{tabular}

\caption{Performance comparison across frontier, fine-tuned (frozen and full), base, and additional models. Each entry is reported as NT/T, where NT denotes models trained without our \texttt{<thinking>} tags and T denotes models trained with them. The NT/T notation applies only to models fine-tuned in this work; scores for all other models are populated under NT. Scores are averaged over three independent evaluation runs.}

\label{tab:results_comparison}
\end{table*}










\vspace{1em}
\subsubsection{Why use only 10,000 videos?}

Although 10,000 training videos are modest compared to large-scale vision-language corpora, our focus is high-quality, grounded, reasoning-focused supervision rather than scale. Each sample undergoes annotation refinement, goal-aware segmentation, and structured reasoning generation, yielding dense, high-signal supervision for grounded and actionable understanding. Our Results (Sec.~\ref{sec:expres-results}) show that this compact yet diverse dataset significantly improves base models. By promoting deliberate visual reasoning, smaller models approach-and in some cases exceed-frontier systems, demonstrating that structured supervision can compensate for scale.

\vspace{1em}
\subsubsection{Why use a 4B-parameter model?}
A central objective of this work is to show that effective video-grounded reasoning does not require frontier-scale architectures. We fine-tune a compact 4B model to demonstrate the impact of structured reasoning in a lightweight setting. Smaller models offer advantages in efficiency, latency, and deployment cost, making them suitable for real-time, edge-based assistance. Despite its size, our reasoning-tuned model achieves competitive performance with frontier systems, confirming that targeted supervision enables grounded behavior without large-scale compute demands.

\subsection{Experimental Results}
\label{sec:expres-results}

\begin{table*}[t]
\centering
\footnotesize
\captionsetup{font=footnotesize, labelfont=bf}
\resizebox{\textwidth}{!}{
\renewcommand{\arraystretch}{1.25}
\begin{tabular}{l l c c c c c c c}
\hline
\textbf{Benchmark} & \textbf{Metric / Dimension} & 
\textbf{Ours FT (4B)} & 
\textbf{Baseline (Qwen3-VL-4B)} & 
\textbf{Qwen3-VL-235B} & 
\textbf{GPT-4o} &
\textbf{FT vs GPT-4o} &
\textbf{GPT-5.2} &
\textbf{FT vs GPT-5.2} \\

\hline

\multirow{4}{*}{TempCompass MC}
& avg\_accuracy 
& 72.0 & 70.1 & 76.1 & 64.1 & \textbf{+7.9} & 77.9 & -5.9 \\
& speed 
& 55.5 & 49.5 & 62.5 & 46.1 & \textbf{+9.4} & 73.5 & -18.0 \\
& order 
& 79.5 & 74.8 & 82.8 & 55.3 & \textbf{+24.2} & 74.8 & \textbf{+4.7} \\
& attribute\_change 
& 79.9 & 78.1 & 83.3 & 76.0 & \textbf{+3.9} & 84.0 & -4.1 \\

\hline

\multirow{6}{*}{EgoThink}
& affordance 
& 65.0 & 52.0 & 47.0 & 67.0 & -2.0 & 76.0 & -11.0 \\
& assistance 
& 47.0 & 40.0 & 46.0 & 46.0 & \textbf{+1.0} & 69.0 & -22.0 \\
& attribute  
& 84.0 & 78.0 & 86.0 & 75.0 & \textbf{+9.0} & 85.0 & -1.0 \\
& situated   
& 84.0 & 79.0 & 89.0 & 70.0 & \textbf{+14.0} & 73.0 & \textbf{+11.0} \\
& location   
& 89.0 & 86.0 & 96.0 & 72.0 & \textbf{+17.0} & 92.0 & -3.0 \\

\hline
\end{tabular}
}
\caption{Results on standard video QA benchmarks comparing our 4B finetuned model with baseline and frontier models. Despite being significantly smaller than frontier models, our finetuned 4B model outperforms GPT-4o on most benchmarks. In TempCompass, our model surpasses GPT 5.2 in \textit{event-order reasoning}. In Egothink, our model surpasses GPT-5.2 in \textit{situated reasoning} and achieves competitive performance in \textit{attribute} and \textit{location reasoning}.}
\label{tab:merged_benchmarks}
\end{table*}



\noindent

\textbf{Evaluating H1.} 
We evaluate whether reasoning-centric fine-tuning improves video-grounded understanding and actionable responses. Table~\ref{tab:results_comparison} compares non-thinking (NT) and thinking (T) variants, where T models are trained with structured \texttt{<thinking>} supervision. A consistent pattern emerges: models trained with the thinking formulation outperform their non-thinking counterparts, particularly in the fine-tuned setting. The best 4B reasoning-tuned model achieves 58.0\%, improving over its NT version (54.67\%) and substantially exceeding its base model. Gains are most evident in goal-oriented evaluation, indicating stronger actionable reasoning.
This improvement indicates that \textit{pause-then-think} supervision encourages the model to more carefully interpret visual evidence before generation. \textit{Thinking} (T) models produce more concise, context-aware responses, whereas NT variants more often rely on generic or loosely grounded outputs. Notably, the reasoning-tuned 4B model approaches-and in some cases rivals-frontier systems, demonstrating that structured reasoning can partially offset scale. A text-only ablation (removing video input) reduces accuracy to 23.7\%, compared to 58.0\% with vision, confirming reliance on visual evidence rather than language priors. Together, these findings support H1: reasoning-centric training measurably improves grounded video understanding and actionable suggestion quality.\\
The \textit{Gemini-robotics-ER-1.5-preview} model did not perform as expected in our evaluation. Its limitations likely stem from difficulty handling unconstrained real-world visual inputs. Official documentation notes sensitivity to poor lighting and reliance on zoomed-in bounding boxes or iterative prompting for accurate perception~\cite{gemini_25}. In contrast, our evaluation uses single-shot, uncropped everyday videos, exposing this fragility. Although Gemini models are also known to hallucinate under ambiguous or out-of-distribution (OOD) inputs, our fine-tuned 4B model demonstrates a stronger generalization of OOD (\ref{ooo}). Without iterative refinement, the frontier model often falls back on textual priors rather than visual grounding, leading to contextual drift and planning failures.

\noindent
\textbf{Evaluating H2.} 
We evaluate whether reasoning-centric fine-tuning enables compact models to match frontier systems without large parameter scaling (Table~\ref{tab:results_comparison}). 
In the thinking (T) setting, the fine-tuned 4B model trained on \textit{Pause-and-think-T} achieves 55.02\% scene accuracy-matching GPT-5.2 (55.13\%) and surpassing GPT-5.0 (52.63\%) and GPT-4o (42.58\%). The Qwen3-4B-VL baseline scores 46.86\%, while larger open models such as InternVL3.5-241B (43.81\%) and Kimi-VL-A3B (36.36\%) perform worse. These results show that structured \textit{pause-and-think} supervision can compensate for scale by improving grounding and deliberate inference.\\ 
Frontier models retain advantages in some goal-planning cases due to strong descriptive priors that produce broadly plausible answers. However, scene-understanding tasks require strict visual grounding, where generic reasoning fails. Here, our reasoning-tuned 4B model excels, producing more specific and context-aligned responses. Overall, H2 is supported: targeted reasoning supervision enables a compact 4B model to rival frontier systems in contextual video QA while improving grounding without large-scale parameter growth.

\noindent
\textbf{Compute efficiency.} 
Figure~\ref{fig:pareto} shows the accuracy--scale tradeoff. Among open-weight models, our fine-tuned 4B model is Pareto-optimal. It achieves 58.0\% accuracy at 59$\times$ fewer parameters than Qwen3-VL-235B (58.9\%), and outperforms GPT-4o (50.3\%), InternVL3.5-241B (50.7\%), and Kimi-VL-A3B (39.3\%) despite being 4--60$\times$ smaller. The model was further optimized using the Ryzen\textsuperscript{TM} AI stack for deployment on a Strix Halo Ryzen AI chip, enabling real-time embedded inference. This efficiency makes it practical for latency-sensitive, edge-based assistance where frontier cloud models are unsuitable.

\textbf{Evaluating H3.}
We evaluate whether structured reasoning supervision improves grounding quality while maintaining concise and polite responses. To measure this, we prompt an external evaluator to score each successful prediction along two dimensions: (1) \textit{conciseness}, defined as how succinctly the agent provides a correct answer without unnecessary verbosity, and (2) \textit{politeness}, defined as whether the response maintains a respectful and user-friendly tone. An aggregate score, termed as \textit{helpfulness} is computed by averaging these two metrics. Results are summarized in Table~\ref{tab:h3_results}. The fine-tuned model achieves the highest aggregate score (0.7601), outperforming both closed (0.7052) and open (0.7072) frontier models. Notably, it demonstrates substantially higher conciseness (0.806) while maintaining competitive politeness (0.7142). In contrast, frontier models exhibit lower conciseness despite comparable politeness levels. \\
It is important to note that politeness is often expressed through the use of courteous phrases (e.g., greetings, acknowledgments, or softeners), which naturally increases response length. Consequently, systems that prioritize politeness often through extensive RLHF alignment \cite{ouyang2022instructgpt} to satisfy human-preferred social norms, may produce more verbose outputs. Because our training objective emphasizes concise, actionable assistance, the model favors brevity while preserving a polite tone rather than maximizing politeness alone. As a result, its politeness score does not surpass frontier models that generate longer responses, but the combined metric indicates a better balance between helpfulness, conciseness, and user-friendly communication.

We also experimented with training variants built on native "thinking" checkpoints\footnote{\url{https://huggingface.co/Qwen/Qwen3-VL-4B-Thinking}} instead of standard instruct models. Contrary to expectations that thinking-oriented models possess stronger reasoning priors, fine-tuning these checkpoints with our dataset did not improve performance over instruct counterparts, and in some cases led to slight degradation (Table~\ref{tab:results_comparison}). This aligns with emerging evidence that VLMs do not consistently benefit from chain-of-thought mechanisms as text-only LLMs do. While structured reasoning can aid logical organization, it may reduce visual grounding in multimodal settings by encouraging reliance on linguistic patterns. In contrast, instruct checkpoints respond more effectively to externally imposed reasoning supervision, yielding stronger grounding and more stable task execution. These results indicate that reasoning gains depend on compatibility between supervision design and multimodal alignment. We therefore adopt instruct checkpoints as the foundation for reasoning-centric fine-tuning.


\begin{table}[t]
\centering
\scriptsize
\captionsetup{font=scriptsize}
\setlength{\tabcolsep}{4pt}
\begin{tabular}{l c c c}
\hline
\textbf{Model} & \textbf{Conciseness} & \textbf{Politeness} & \textbf{Agg. Helpfulness} \\
\hline
\rowcolor{frontier}
GPT 5.2 & 0.634 & 0.777 & 0.705 \\
\rowcolor{lightpurple}
Qwen3VL-235B & 0.658 & 0.757 & 0.707 \\
\rowcolor{finetuned}
Ours & \textbf{0.806} & 0.714 & \textbf{0.760} \\
\hline
\end{tabular}
\caption{Conciseness and politeness evaluation. Conciseness measures how succinctly the agent provides a correct answer without unnecessary verbosity. Politeness evaluates whether the response maintains a respectful and user-friendly tone. The aggregate score is the average of the two metrics.}
\label{tab:h3_results}
\end{table}


\noindent

\subsubsection{Generalization and Targeted Skill Transfer}
\label{ooo}
Table~\ref{tab:merged_benchmarks} demonstrates strong generalization of our compact training setup. Although the model was not trained on these benchmarks, the 4B fine-tuned variant consistently outperforms its baseline across all evaluated metrics.
Despite its small size, the model remains competitive with much larger frontier systems, particularly on capabilities targeted by our dataset. On TempCompass, the fine-tuned model surpasses GPT-4o on all reported metrics, including a +24.2\% improvement in temporal \textit{order} reasoning and a +7.9\% gain in overall accuracy. It also exceeds GPT-5.2 in event-\textit{order} reasoning by +4.7\%. A similar trend appears on EgoThink, where the model outperforms GPT-4o in four of five dimensions, including \textit{location} (+17.0\%), \textit{situated} reasoning (+14.0\%), \textit{attribute} recognition (+9.0\%), and \textit{assistance} (+1.0\%). It also surpasses GPT-5.2 in \textit{situated} reasoning (+11.0\%) and remains competitive in \textit{attribute} (-1.0\%) and \textit{location} (-3.0\%). Overall, structured reasoning supervision from $\sim$10,000 curated videos transfers beyond the training distribution. These targeted gains-particularly in spatial grounding and temporal order reasoning-demonstrate that a compact, high-signal dataset can induce advanced capabilities without large-scale parameter growth or benchmark-specific training.

\vspace{1em}
\subsubsection{Qualitative Behavioral Analysis}
Figure \ref{fig:example} highlight key behavioral differences between the reasoning-tuned model and frontier systems. In our qualitative experiments we observed the following: First, the fine-tuned model shows stronger spatial and directional grounding, resolving left--right orientation, object-relative positioning, and action direction (e.g., clockwise vs. anticlockwise) more consistently. Frontier models occasionally produce directionally inconsistent or loosely grounded instructions, indicating weaker perspective-aware reasoning. Second, structured \textit{pause-and-think} supervision enables task-aware inference under partial observability. When objects are occluded or unclear, the model leverages task context and goal progression to infer plausible actions instead of hallucinating unrelated objects. Frontier models more often rely on linguistic priors, producing visually unsupported responses. Third, the reasoning-centric model is more robust to low-quality input (blur, reduced resolution, motion). It maintains coherent, task-consistent outputs, suggesting improved high-level contextual integration rather than reliance on fine visual details. Overall, \textit{pause-and-think} supervision improves spatial grounding, contextual inference, and robustness under imperfect visual conditions.

\section{Conclusions and Future Works}

We presented a reasoning-centric training framework for generative assistive action suggestion from video, shifting beyond recognition-based QA toward concise, grounded, actionable guidance. By supervising structured reasoning over temporally aligned video evidence, the dataset encourages models to think before responding, reducing hallucinations and improving contextual fidelity. Experiments show that targeted reasoning supervision enables a compact $\sim$4B model to compete with frontier systems while remaining compute-efficient and edge-deployable. Strong gains on assistive and sequential reasoning tasks confirm that a high-signal, compact dataset can improve performance without direct benchmark exposure. These findings demonstrate that data quality and reasoning supervision-not parameter scaling alone-drive effective video-grounded assistance.\\
Future work will explore richer spatial representations and persistent memory to improve long-horizon grounding and task continuity. Additionally, we plan to investigate building Vision-Language-Action (VLA) models inspired by our video-grounded assistive guidance, enabling robots to translate contextual video understanding into physically grounded actions. Such models could enhance VLA capabilities in dynamic environments involving human-robot collaboration. A key next step is deploying our reasoning-tuned model on a physical embodied agent or robot, closing the perception-reasoning-action loop in real-world settings.


\balance

\bibliographystyle{IEEEtranS}
\bibliography{references}

\end{document}